\documentclass[conference]{IEEEtran}
\IEEEoverridecommandlockouts
\usepackage{cite}
\usepackage{amsmath,amssymb,amsfonts}
\usepackage{algorithmic}
\usepackage{graphicx}
\usepackage{textcomp}
\usepackage{xcolor}
\def\BibTeX{{\rm B\kern-.05em{\sc i\kern-.025em b}\kern-.08em
    T\kern-.1667em\lower.7ex\hbox{E}\kern-.125emX}}
\begin{document}

\title{Multiscale Voxel Based Decoding For Enhanced Natural Image Reconstruction From Brain Activity \\
\thanks{*Corresponding author}
}

\makeatletter
\newcommand{\linebreakand}{%
  \end{@IEEEauthorhalign}
  \hfill\mbox{}\par
  \mbox{}\hfill\begin{@IEEEauthorhalign}
}
\makeatother

\author{
  \IEEEauthorblockN{Mali Halac*}
   
  \IEEEauthorblockA{\textit{Electrical and Computer Eng.} \\
    \textit{Drexel University}\\
    Philadelphia, PA \\
    mh3636@drexel.edu}
   
  \and
  \IEEEauthorblockN{Murat Isik}
  \IEEEauthorblockA{\textit{Electrical and Computer Eng.} \\
    \textit{Drexel University}\\
    Philadelphia, PA \\
    mci38@drexel.edu}
  \and
  \IEEEauthorblockN{Hasan Ayaz}
  \IEEEauthorblockA{\textit{School of Biomedical Eng.} \\
    \textit{Drexel University}\\
    Philadelphia, PA \\
    ha45@drexel.edu}
  \linebreakand 
  \IEEEauthorblockN{Anup Das}
  \IEEEauthorblockA{\textit{Electrical and Computer Eng.} \\
    \textit{Drexel University}\\
    Philadelphia, PA \\
    ad3639@drexel.edu}
}

\maketitle

\begin{abstract}
Reconstructing perceived images from human brain activity monitored by functional magnetic resonance imaging (fMRI) is hard, especially for natural images. Existing methods often result in blurry and unintelligible reconstructions with low fidelity. In this study, we present a novel approach for enhanced image reconstruction, in which existing methods for object decoding and image reconstruction are merged together. This is achieved by conditioning the reconstructed image to its decoded image category using a class-conditional generative adversarial network and neural style transfer. The results indicate that our approach improves the semantic similarity of the reconstructed images and can be used as a general framework for enhanced image reconstruction.
\end{abstract}


\section{Introduction}
Understanding the representation of sensory information in the human brain has long been a topic of interest in neuroscience\cite{i-a1}. Successful decoding of the visual stimuli from brain activity may pave the way to groundbreaking technologies. Some futuristic applications range from being able to record and play back every moment of our lives to revealing the contents of mental imagery. 

Earlier studies have demonstrated that the neural activation patterns obtained by functional magnetic resonance imaging (fMRI) are informative about the visual stimuli\cite{i1,i2,i3,i4}. Integration of deep neural networks (DNN) to the field of neuroscience have further improved the visual decoding task\cite{i5,i6}. In one such study, Horikawa and Kamitani developed a deep convolutional neural network that predicts the image features from the fMRI voxels using a set of linear regression algorithms\cite{i7}. Their model can predict the features for images whose categories were not used for training. They show that the novel image categories can be identified by comparing the predicted features with the category average image features. In another study, Qiao et al. proposed a bidirectional recurrent neural network (BRNN) to decode image categories\cite{i-a2}. Their model makes use of the bidirectional information flows as well as the hierarchical visual features. Recently, Akamatsu et al. have shown that a better category decoding accuracy can be obtained by extracting the shared information from multi-subject fMRI recordings\cite{i-a3}.

A relatively harder problem, image reconstruction, requires an invertible mapping between the visual stimuli and the fMRI voxels. Shen et al. has shown that such a mapping can be formed by combining hierarchical neural representations using a deep network\cite{i8}. In another study, Beliy et al. proposed a self-supervised reconstruction model that allows training on unlabeled fMRI data\cite{i9}. In a later study, Ren et al. explored adversarial representation learning with knowledge distillation for improved image reconstruction\cite{i10}. Although these methods were successful in capturing the general shape and color, the resulting images were blurry, unintelligible and they suffered from low fidelity. Recently, Qiao et al. developed a GAN based Bayesian model that produces grey scale reconstructions\cite{i11}. Their model generates images with improved fidelity and naturalness by taking the image categories into account. 

In this study, we propose a novel framework that can be employed with most image reconstruction and category decoding methods to enhance the reconstructed images. We expand on the idea of exploiting the decoded category of the perceived image to enhance the reconstructed image. This merging of two different techniques is achieved by using a class-conditional generative adversarial network as well as a neural style transfer. The class-conditional generator allows to increase the structural similarity whereas the style transfer preserves the pixel-wise relationship resulting in an improved fidelity. The results suggest that our framework can be adopted as a general method to enhance images reconstructed from brain activity.

\section{Background}
\subsection{Generative Adversarial Network}
First introduced by Goodfellow et al., generative adversarial networks (GAN)\cite{b1} have seen wide adaptations in many fields involving image and audio processing such as biomedical informatics\cite{b2}. A generative adversarial network consists of a set of two neural networks---generator and discriminator. The goal of the generator is to fool the discriminator by mapping from a latent space to the real space (i.e. pixel space). The discriminator, on the other hand, learns to distinguish synthesized from real data. Thus, a generative adversarial network can be seen as a zero sum game between the generator and the discriminator:
\begin{equation}
    \begin{split}
        \min\limits_G \max\limits_D V(D,G) = \min_{G}\max_{D}\mathbb{E}_{x\sim p_{\text{data}}(x)}[\log{D(x)}] \\
        + \mathbb{E}_{z\sim p_{\text{generated}}(z)}[1 - \log{D(G(z))}]
    \end{split}
    \label{eq:GAN}
\end{equation}
where $x$ is the real data, $z$ is the latent representation of the synthesized data, $G$ is the generator and $D$ is the discriminator\cite{b1}. 

A variant of GAN called the conditional generative adversarial network (CGAN) allows control over the generator\cite{b3}. The conditioning can be performed based on the class labels by modifying the GAN equation (Eq. \ref{eq:GAN}).
\begin{equation}
    \begin{split}
        \min\limits_G \max\limits_D V(D,G) = \min_{G}\max_{D}\mathbb{E}_{x\sim p_{\text{data}}(x)}[\log{D(x|y)}] \\
        + \mathbb{E}_{z\sim p_{\text{generated}}(z)}[1 - \log{D(G(z|y))}]
    \end{split}
    \label{eq:CGAN}
\end{equation}
where the inputs to the discriminator and generator are changed by the conditional probability of real data x given a condition y and the conditional probability of synthesized data z given the condition y.

\subsection{Neural Style Transfer}
Neural style transfer was first introduced by Gatys et al. as a computerized mean to imitate human's innate ability to distinguish between the content and style of an image\cite{b4}. Through the use of convolutional neural networks (CNN), they showed that it is possible to extract the content and style information from images. Moreover, the extracted content and style can be recombined using CNN feature activations\cite{b5}. 
Neural style transfer is mostly used for artistic purposes in which the content of a natural image is stylized by the style of an artwork.

\section{Methods}
\subsection{Data Set}
We made use of the preprocessed fMRI data set shared by Kamitani Lab at Kyoto University\cite{m1}. The data set is distributed under Creative Commons License. The fMRI data was collected while 5 healthy subjects were presented with natural images from 200 categories from the ImageNet database. The fMRI signals of V1, V2, V3, V4, LOC, FFA and PPA were recorded during an image presentation experiment. The training set consists of 8 images from 150 categories, a total of 1,200 images. The test set consists of 1 image from 50 categories that were not contained in the training set. Each image in the training set is presented only once, whereas each image in the test set was presented 35 times. The test set recording is obtained by averaging across all 35 trials.

\subsection{Overview of the Proposed Framework}
\begin{figure}[h!]
   \centering   
   \includegraphics[scale=0.3]{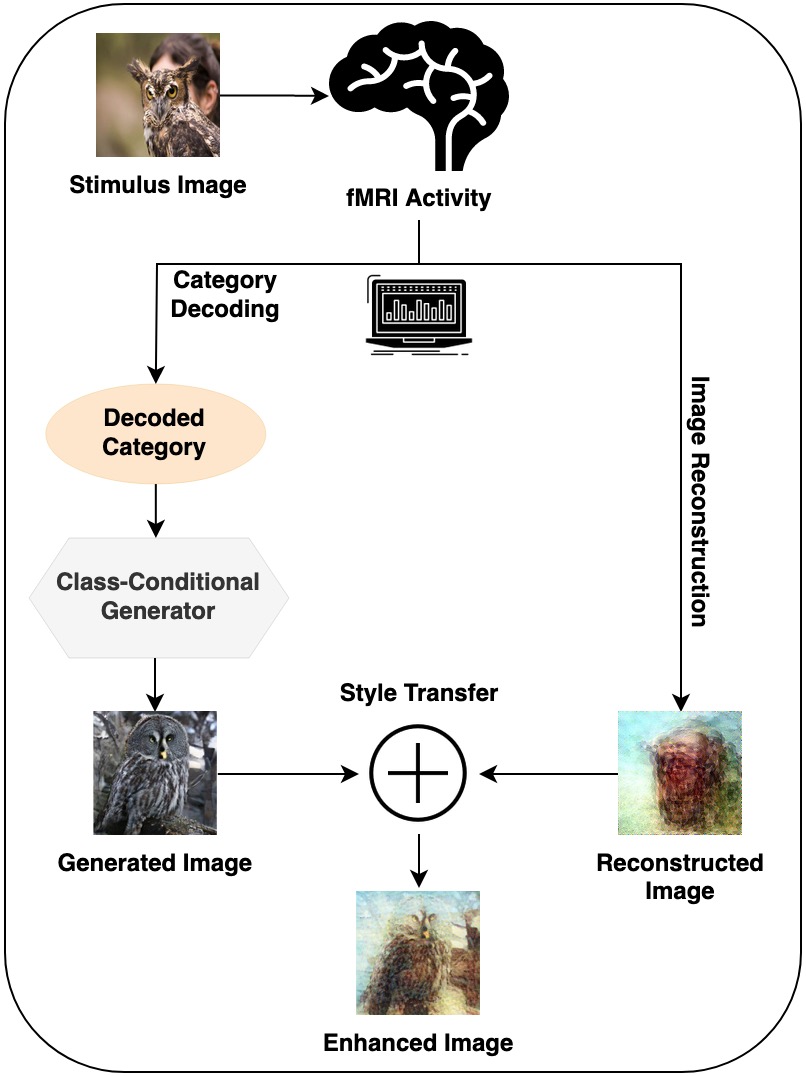}
   \caption{{\bf The Proposed Framework.} The recorded fMRI activity corresponding to the stimulus image is forwarded to category decoding and image reconstruction algorithms. The decoded category of the perceived image is used to generate a class-conditional image. Then, the contents of the generated image is merged with the style of the reconstructed image to obtain the enhanced image.}
   \label{proposed_framework}
\end{figure}

The proposed framework, illustrated in Fig. \ref{proposed_framework}, consists of three main parts: category decoding, image reconstruction and style transfer. For a given fMRI activity, the category of the stimulus image is decoded using the algorithm of choice. The decoded category is then fed into the pre-trained class-conditional generator of BigGAN\cite{m2} which generates a natural image according to the decoded category. The second part of our framework tries to reconstruct the stimulus image based on the fMRI activity. The specific algorithm to use for either category decoding or image reconstruction is up to the researcher. The reconstructed image often captures the general shape and color of the perceived image; however, it lacks the structural content that makes it informative about the contents of the image. Thus, a style transfer is applied using the contents of the generated image and the style of the reconstructed image.

\begin{figure}[h!]
   \centering   
   \includegraphics[scale=0.3]{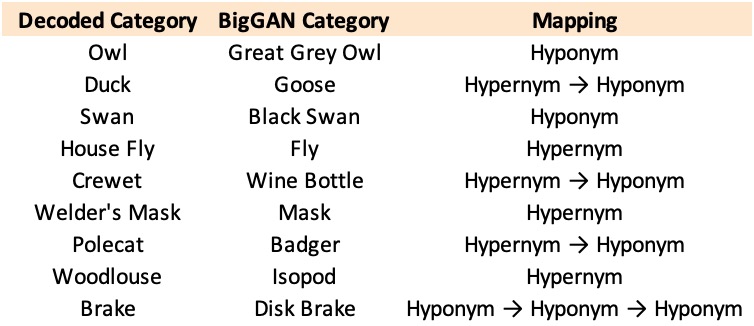}
   \caption{{\bf Category Mapping.} The decoded categories that did not exist in the pre-trained BigGAN were manually mapped to BigGAN categories. The mapping was performed by taking the closest hyponym or hypernym available in BigGAN categories. The first column shows the decoded category. The second column shows the corresponding BigGAN category. The third column shows the mapping order. Hypernym $\rightarrow$ Hyponym means that the decoded category's hypernym of hyponym was taken to arrive to the closest BigGAN category.}
   \label{category_mapping}
\end{figure}

\begin{figure*}[h!]
   \centering   
   \includegraphics[scale=0.21]{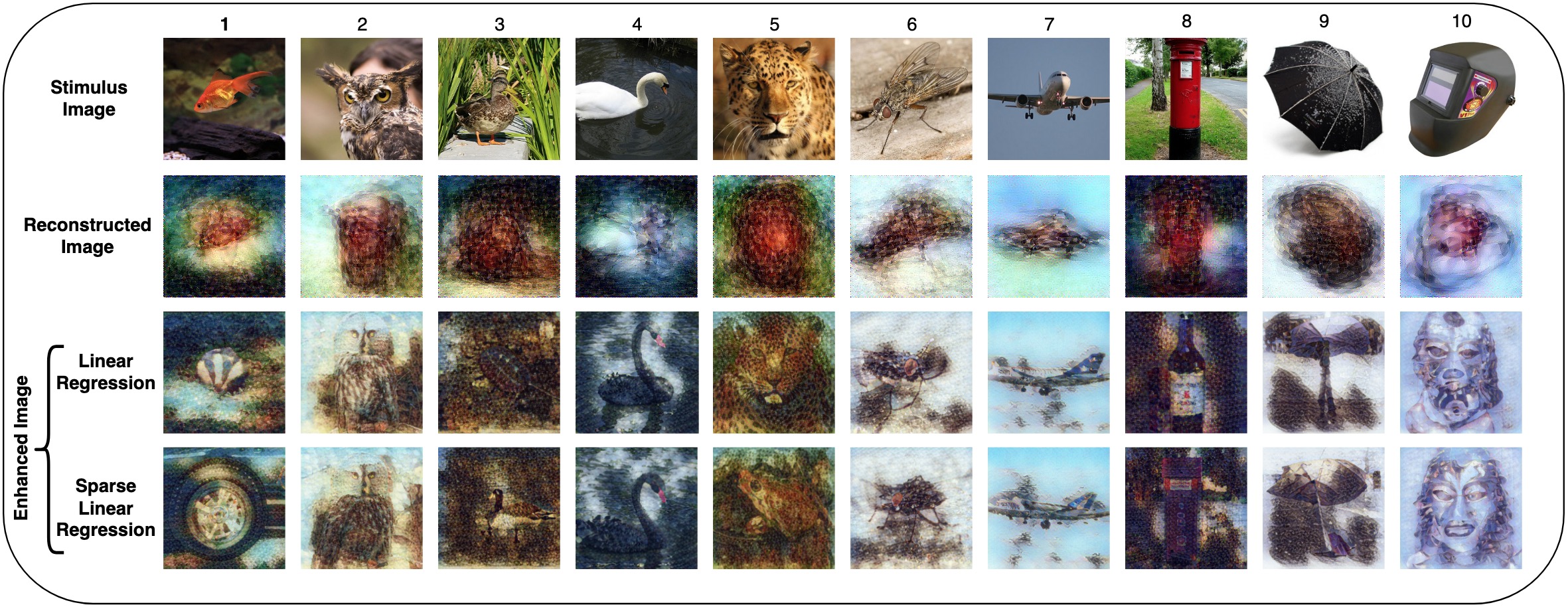}
   \caption{{\bf Effect of Category Decoding.} Two different category decoding methods and resulting enhanced images are shown. The first row shows the stimulus images. The second row corresponds to the reconstructions obtained by Shen et al.'s image reconstruction method\cite{i8}. The last two rows show the enhanced images whose categories were predicted by using Horikawa and Kamitani's linear regression and sparse linear regression algorithms\cite{i7}. Each image is numbered at the top. Those numbers correspond to the indices for images in Fig. \ref{category_decoding_table}.}
   \label{category_decoding_fig}
\end{figure*}

\begin{figure*}[h!]
   \centering   
   \includegraphics[scale=0.21]{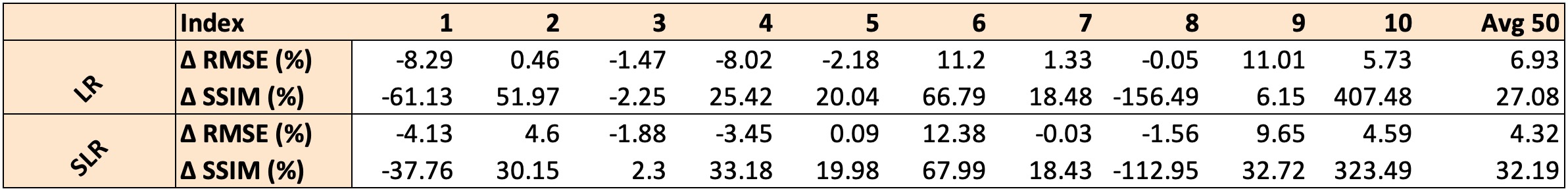}
   \caption{{\bf Category Decoding Evaluations.} The resulting images shown in Fig. \ref{category_decoding_fig} are evaluated using the root mean square error (RMSE) and structural similarity (SSIM) index. The table shows the percent change in RMSE and SSIM values between the reconstructed image and the enhanced image. LR stands for the linear regression algorithm whereas SLR stands for sparse linear regression algorithm which were used to decode the image categories. The column labeled "Avg 50" shows the average values across 50 test samples.}
   \label{category_decoding_table}
\end{figure*}

In this study, we used Horikawa and Kamitani's generic object decoding model which rely on linear regression and sparse linear regression algorithms\cite{i7}. The brain ROIs used for category decoding are V1-V4, LOC, FFA, PPA. For each ROI, a total of 500 voxels are selected based on the correlation with the target in the training session. The category decoding was performed by computing the Pearson's correlation coefficients between the decoded and category-average feature vectors in the candidate set. The candidate set consists of 29 randomly selected categories and the true category, total of 30 categories. While Horikawa and Kamitani's model can predict 15,372 categories in ImageNet\cite{m3}, the pre-trained class-conditional generator of BigGAN has only 1,000 categories. Therefore, we manually mapped the decoded categories to the BigGAN categories by looking at their closest hyponyms and hypernyms (Fig.\ref{category_mapping}).

\subsection{Evaluation Metrics}
The quality of enhanced images are evaluated using root mean square error (RMSE) and structural similarity index (SSIM). RMSE is a useful metric to determine the pixel-wise relationship between two images as well as the amount of deviation from the actual pixel values. RMSE of two images x and y is calculated as follows:
\begin{equation}
    RMSE = \sqrt{\frac{1}{MN}\sum_{i=0}^{M-1}\sum_{j=0}^{N-1}(y_{ij} - x_{ij})^{2}}
    \label{eq:RMSE}
\end{equation} 
where M and N are the height and width of the images, y\textsubscript{ij} the stimulus image and x\textsubscript{ij} enhanced/reconstructed image. The lowest value for RMSE is 0 which indicates maximum pixel-wise similarity. SSIM, on the other hand, is useful to determine the perceived image quality as it is correlated with the quality perception of the human visual system\cite{m4,m5}. It measures the structural similarity between two images x and y according to the following formula:
\begin{equation}
  SSIM(x,y) = \frac{(2\mu_x\mu_y + C_1) + (2 \sigma _{xy} + C_2)} 
    {(\mu_x^2 + \mu_y^2+C_1) (\sigma_x^2 + \sigma_y^2+C_2)}
  \label{eq:SSIM}
\end{equation}
where $\mu_x$ and $\mu_y$ are the averages of x and y, $\sigma_x^2$ and $\sigma_y^2$ are the variances of x and y, $\sigma_{xy}$ is the covariance of x and y and $C_1$ and $C_2$ are constants that ensure stability. SSIM exists in the range [-1,1] with 1 indicating maximum similarity. Additionally, we have performed two n-way identification tasks with n=2,5,10 for better comparison with previous methods. For each reconstructed and enhanced image, the task is to find the ground truth image among a candidate of n images, one being the ground truth while the rest were randomly selected. The identification was performed by computing either the Pearson correlation or SSIM score between the enhanced \& candidate image and reconstructed \& candidate image. The candidate image with the max score was determined to be the ground truth. The n-way identification task was repeated 50 times and averaged over the reconstructed \& enhanced images.

For the subjective assessment, two behavioral experiments were conducted with a group of 5 raters (2 males and 3 females, aged between 21 and 30 years). In the first experiment, the raters were shown the reconstructed images and asked to identify the image category. The same procedure was repeated for the enhanced images as well. In the second experiment, the raters viewed a display presenting the stimulus image at the top, the reconstructed image at the bottom left and the enhanced image at the bottom right. They were then asked to evaluate the similarity of the reconstructed image and the enhanced image to the stimulus image on a scale of 0-10 with 0 representing not similar and 10 representing the same. 

\section{Results}
\begin{figure*}[t!]
   \centering   
   \includegraphics[scale=0.41]{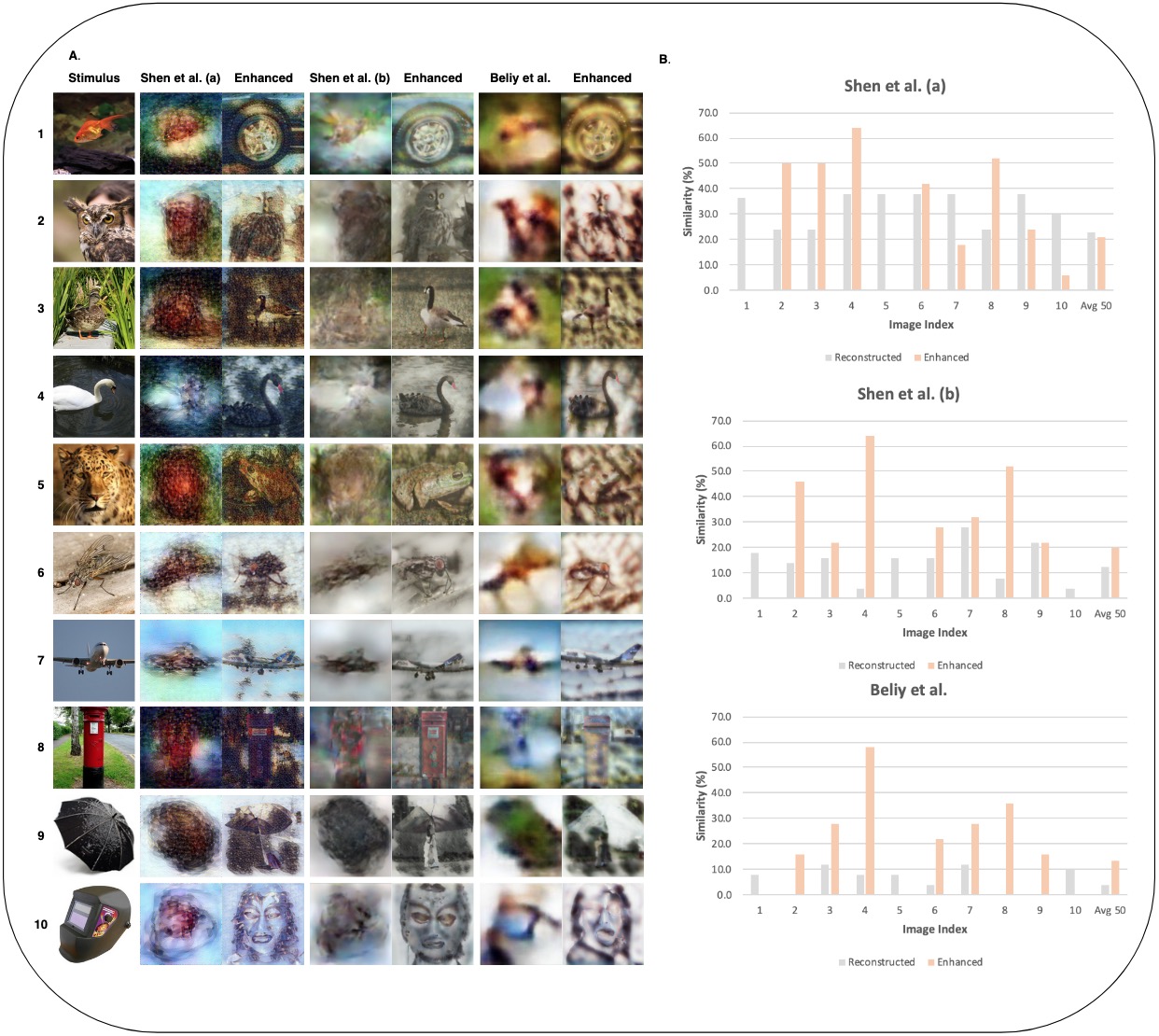}
   \caption{{\bf Enhanced Images.} {\bf Part A} shows the resulting images obtained by various image reconstruction algorithms. The columns "Shen et al. (a)"\cite{i8}, "Shen et al. (b)"\cite{r1} and "Beliy et al."\cite{i9} indicate the image reconstruction algorithms used. On their right column are their corresponding enhanced images. Horikawa and Kamitani's sparse linear regression algorithm is used for the category decoding \cite{i7}. {\bf Part B} shows the results for the subjective assessment. Each plot displays the similarity of reconstructed and enhanced images to the stimulus image. "Avg 50" shows the average values across 50 test samples.}
   \label{image_reconstruction_fig}
\end{figure*}

\begin{figure*}[t!]
   \centering   
   \includegraphics[scale=0.22]{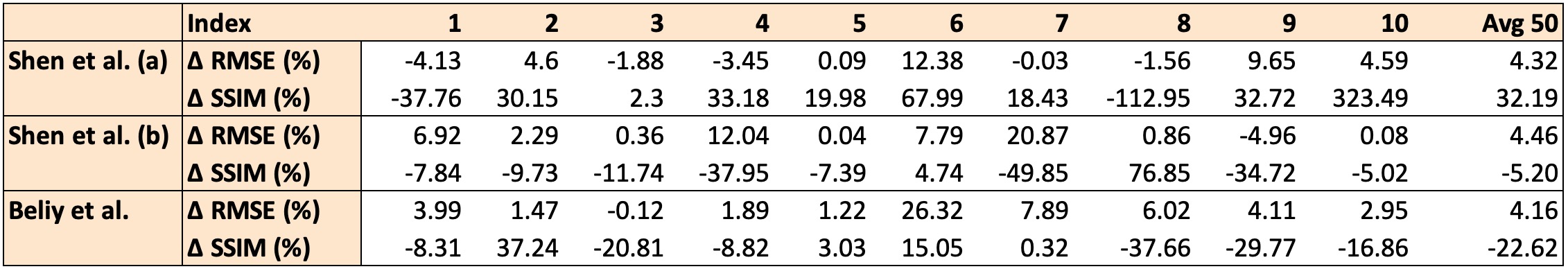}
   \caption{{\bf Evaluation of Enhanced Images.} The enhanced images shown in Fig. \ref{image_reconstruction_fig} are evaluated using root mean square error (RMSE) and structural similarity (SSIM) index. The first column shows the image reconstruction algorithm used to obtain the enhanced images. The values in the table are obtained by computing the percent change in RMSE and SSIM. The column labeled "Avg 50" shows the average values across 50 test samples.}
   \label{image_reconstruction_table}
\end{figure*}

We first tested our framework with two different category decoding algorithms to see the effect of category decoding on the resulting enhanced images. Fig. \ref{category_decoding_fig} illustrates the stimulus images, reconstructed images obtained by Shen et al.'s image reconstruction model \cite{i8} and the enhanced images. The algorithms used for category decoding, noted as linear regression and sparse linear regression on Fig. \ref{category_decoding_fig}, were proposed by Horikawa and Kamitani\cite{i7}. The evaluations of the resulting images can be seen in Fig. \ref{category_decoding_table}. For both the linear regression and sparse linear regression algorithms, the enhanced images prove to have an improved structural similarity to the stimulus images given that the predicted category was correct. The enhanced images for which the category prediction was wrong (image 1 LR-SLR, 3 LR, 8 LR) have the wrong semantic content resulting in a decrease in structural similarity (Figs. \ref{category_decoding_fig}-\ref{category_decoding_table}). 

In the methods section, it is mentioned that the decoded categories were manually mapped to the BigGAN categories by using their hyponym or hypernym. The decoded category for image 4 in Fig. \ref{category_decoding_fig}, for example, was "Swan" for both the linear regression and sparse linear regression algorithms. This decoded category was mapped to its BigGAN category by taking its hyponym which corresponds to "Black Swan" (Fig. \ref{category_mapping}). Consequently, the enhanced images for image 4 shows an improvement in structural similarity: +25.42\% and +33.18\% respectively for linear regression and sparse linear regression algorithms. However, as can be noticed from Fig. \ref{category_decoding_fig}, the stimulus image belongs to a "White Swan", whereas the enhanced images show a "Black Swan". Similarly, the decoded category for image 3 with sparse linear regression algorithm was "Duck", whereas its corresponding BigGAN category was "Goose" (Fig. \ref{category_mapping}). This resulted in a slight increase in the structural similarity.  A similar case is observed for image 10 whose decoded category was mapped to its closest BigGan category by taking its hypernym (Fig. \ref{category_mapping}). Although, the stimulus image shows a "Welder's Mask", the enhanced images belong to a "Mask" which is a more general category. Fig. \ref{category_decoding_table} shows that the enhanced versions of image 10 have significantly improved structural similarities despite the difference in the semantic levels between the true category (Welder's Mask) and the predicted category (Mask). These results show the importance of category decoding on the enhanced images.

We then evaluated our framework by using 3 recently proposed image reconstruction methods. Fig. \ref{image_reconstruction_fig} Part A illustrates the obtained images and Part B shows the results of the subjective assessment. The experimental details of the subjective assessment can be found in the methods section. Fig. \ref{image_reconstruction_table} shows the percent change in RMSE and SSIM values for each image reconstruction method. The subjective assessment results indicate that no similarity was detected in the enhanced images number 1 and 5. This was expected since the categories of these images were wrongly decoded. Similarly, no improvement in similarity is observed for image 10. That is because of the manual mapping between the decoded category (Welder's Mask) and the BigGan category (Mask). For the rest of the images, the subjective assessment indicates an improved similarity for the enhanced images.

The slight changes in the RMSE values (mean = 4.32, 4.46, 4.16\%; standard deviation = 5.8, 7.2, 7.9\%) observed from Fig. \ref{image_reconstruction_table} proves that the change in the pixel-wise similarity is minimized.

We have also performed two n-way identification tasks with n=2,5,10. Fig. \ref{nway_identification} illustrates n-way identification results based on Pearson correlation. In the 2-way experiment, our method was outperformed by a mean identification accuracy of 20.6\%. In the 5- and 10-way experiments, however, our enhanced images catch up with the identification accuracy of the reconstructed images and in some cases outperform them. One observation consistent across all Pearson correlation based n-way experiments is that the Beliy et al.'s reconstruction method outperforms ours. Fig. \ref{nway_identification_ssim}, on the other hand, shows the n-way identification results based on the SSIM. It can be observed from Fig. \ref{nway_identification_ssim} that while the previous reconstruction methods show better identification accuracies in the 2-way experiment; our method shines as the difficulty level increases. In the 5- and 10-way experiments, enhanced images are shown to outperform previous reconstruction methods. Consistent with the previous results, these findings suggest that our method increases the structural similarity while preserving the pixel-wise similarity.

\begin{figure}[h!]
   \centering   
   \includegraphics[scale=0.19]{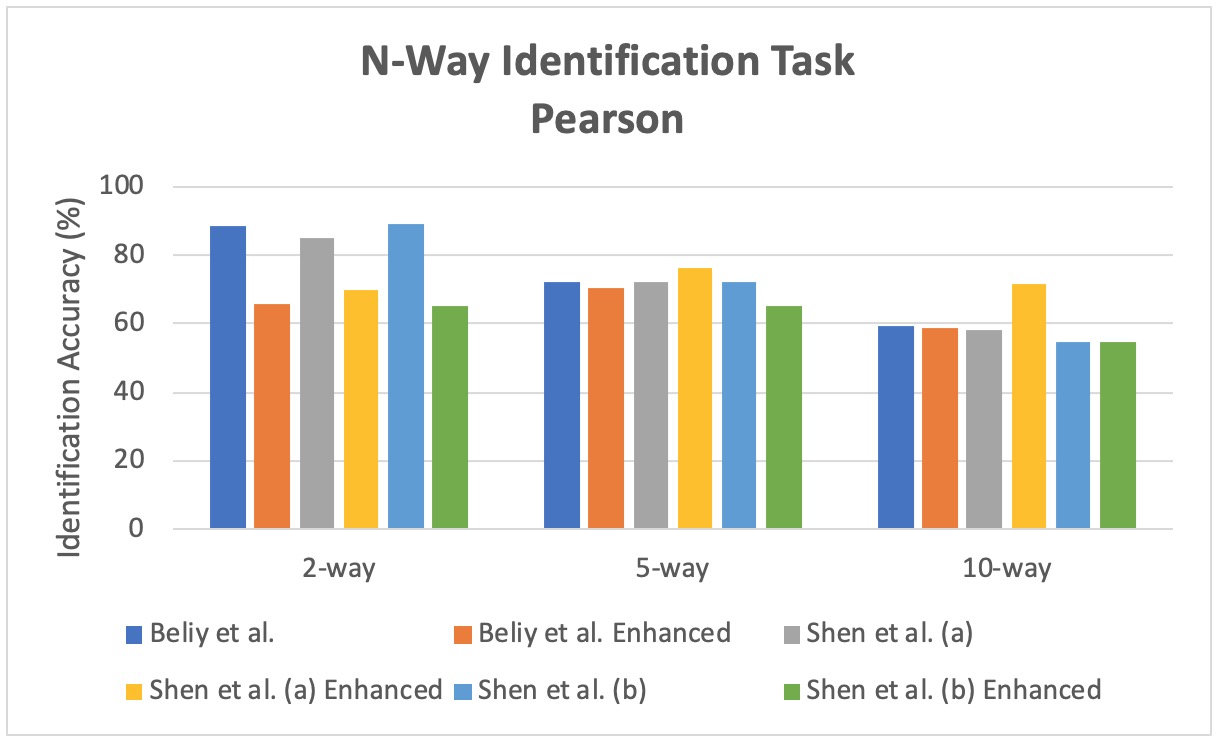}
   \caption{{\bf N-way Identification Task: Pearson Correlation.} N-way identification task performed for n=2,5,10. The Pearson correlations of each reconstructed and enhanced image were computed and averaged over 50 runs.}
   \label{nway_identification}
\end{figure}

\begin{figure}[b!]
   \centering   
   \includegraphics[scale=0.19]{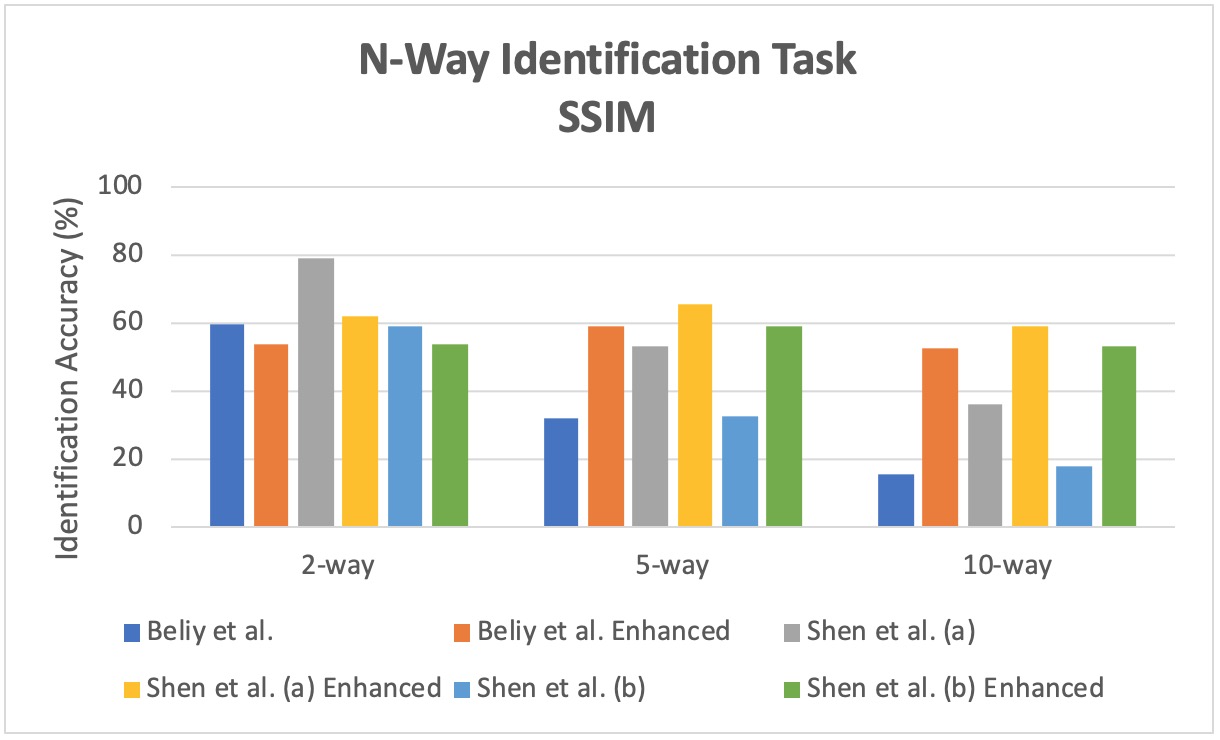}
   \caption{{\bf N-way Identification Task: SSIM.} N-way identification task performed for n=2,5,10. The SSIM of each reconstructed and enhanced image were computed and averaged over 50 runs.}
   \label{nway_identification_ssim}
\end{figure}

\begin{figure}[h!]
   \centering   
   \includegraphics[scale=0.2]{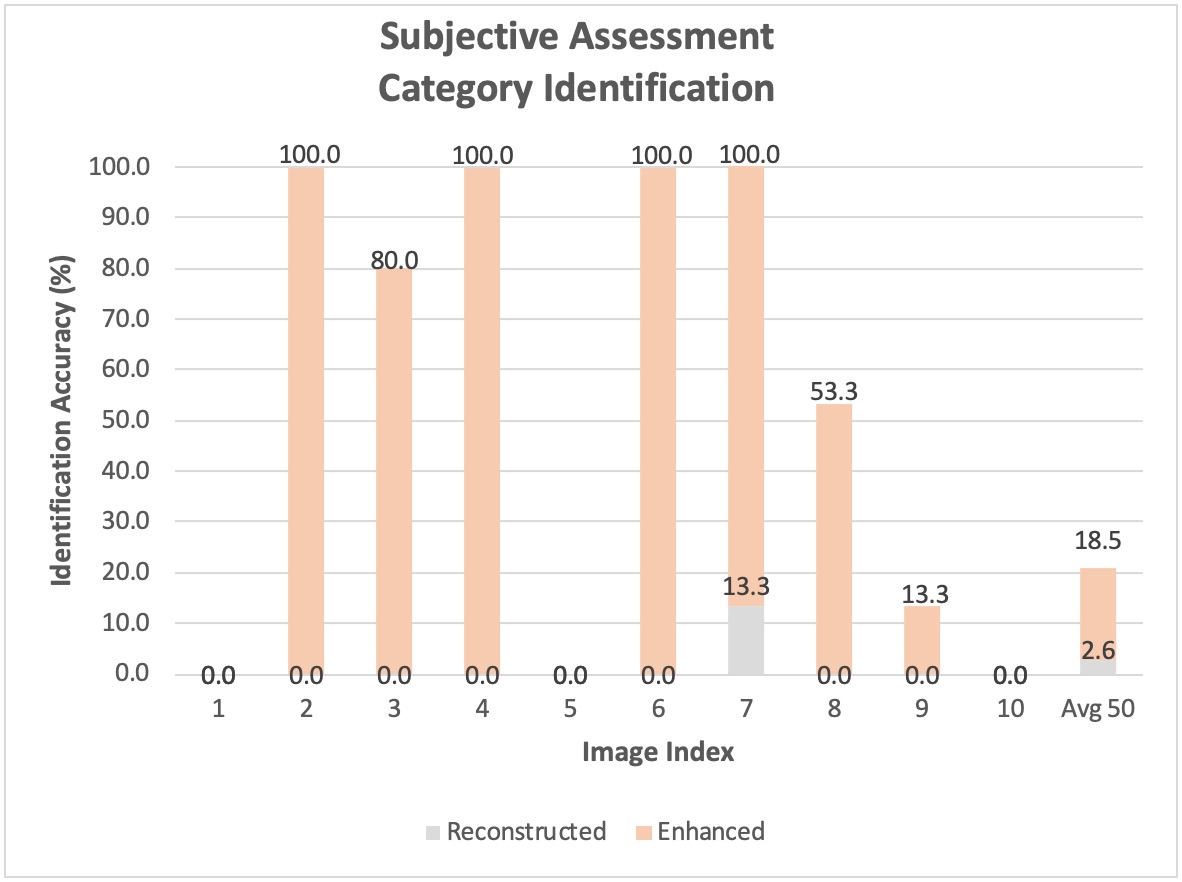}
   \caption{{\bf Category Identification.} Subjective assessment evaluation results for category identification experiment. The participants were shown the reconstructed and enhanced images and asked to identify the category of the presented image. The y-axis shows the identification accuracy which was obtained by averaging across the 5 raters. The x-axis corresponds to the image index numbers. "Avg 50" shows the average values across 50 test samples.}
   \label{category_identification}
\end{figure}

The performance of our framework was further evaluated using a second subjective assessment where the participants were asked to identify the category of the reconstructed and enhanced images. Fig. \ref{category_identification} illustrates the identification accuracy for each image used in our experiments. The mean identification accuracy across all 50 test samples for reconstructed images is 2.6\%, compared to 18.5\% for the enhanced images. It is important to note that images whose decoded categories were incorrect (e.g. images 1 and 5) decrease the mean identification accuracy. The recorded mean identification accuracy for the enhanced images whose decoded categories were correct is 68.3\%.

\section{Discussion}
In this study, we have presented a novel approach to enhance natural images reconstructed from human fMRI activity. We have shown that it is possible to decrease the ambiguity in the reconstructed image by conditioning it to its decoded category. This way, the semantic content of the reconstructed image is maximized, making it more informative about the categorical structures. 

The resulting enhanced images are shown to have an overall increased similarity to the stimulus images. One dis/advantage of our method is that it conditions the reconstructed image to the decoded category. Thus, the enhanced image is highly dependent on the correct decoding of the image category. It can be observed from the samples 3,7,9 in Fig. \ref{category_decoding_fig} that even when the object category is correctly decoded, the orientation of the BigGAN generated image may cause a strong distortion in the enhanced image. In crime scenes, for example, the orientation of a weapon may greatly affect the outcome of an investigation. Similarly, while turning a white swan into a black swan may not be catastrophic in some contexts, they are still quite different from each other. Plus, in some cultural settings changing an individual's color may be problematic. These results show that it is crucial to decode a richer set of features that can be used together with the class label to generate samples that are more aligned with the input stimulus. 

The pre-trained class-conditional generator of BigGAN required a mapping between the decoded categories (15,372) and the BigGAN categories (1,000). The difference in the hierarchy of some of the decoded and BigGan categories are shown to result in a decreased similarity. Thus, for future applications, a class-conditional generator with a larger number of classes may give better results.

The main advantage of our method is that it can be adapted to different algorithms and models, thus potentially allowing to improve system performance by just replacing single modules such as the decoding classifier.
\section*{Acknowledgment}

This material is based upon work supported by the U.S. Department of Energy under Award Number DE-SC0022014 and by the National Science Foundation under Grant Nos. CCF-1942697 and CCF-1937419.

\vspace{12pt}

\end{document}